# Design and Implementation of Global Path Planning System for Unmanned Surface Vehicle among Multiple Task Points

## Yanlong Wang*

School of Automation Science and Electrical Engineering,
Beihang University,
XueYuan Road No.37,
HaiDian District, Beijing, China
Email: wylloong@163.com
*Corresponding author

## Xuemin Yu

Institute of Information Engineering,
Chinese Academy of Sciences,
YuQuan Road No.19,
ShiJingShan District, BeingJing, China
Email: yuxuemin@iie.ac.cn

## Xu Liang

School of Automation Science and Electrical Engineering,
Beihang University,
XueYuan Road No.37,
HaiDian District, Beijing, China
Email: zy1503245@buaa.edu.cn

**Abstract:** Global path planning is the key technology in the design of unmanned surface vehicles. This paper establishes global environment modelling based on electronic charts and hexagonal grids which are proved to be better than square grids in validity, safety and rapidity. Besides, we introduce Cube coordinate system to simplify hexagonal algorithms. Furthermore, we propose an improved $A^*$ algorithm to realize the path planning between two points. Based on that, we build the global path planning modelling for multiple task points and present an improved ant colony optimization to realize it accurately. The simulation results show that the global path planning system can plan an optimal path to tour multiple task points safely and quickly, which is superior to traditional methods in safety, rapidity and path length. Besides, the planned path can directly apply to actual applications of USVs.

University, Shanghai, China, in 2015 and the MS degree from Beihang University, Beijing, China, in 2018. After he completed his MS, he joined DiDi as an Algorithm Engineer in Autonomous Vehicle Group. His research interests include path planning, collision avoidance, simulation system, intelligent control.

Xuemin Yu is currently a master student in Institute of Information Engineering, University of Chinese Academy of Sciences, Beijing, China. She received the BS degree from Communication University of China, Beijing, China, in 2015. Her research interests include big data, intelligent decision, network security.

Xu Liang is currently an associate professor in School of Automation Science and Electrical Engineering, Beihang University, Beijing, China. He received the BS degree from Beijing Institute of Technology, Beijing, China, the MS degree and the PhD degree from Beihang University, Beijing, China. His research interests include flight vehicle design, detection technology and automatic equipment, intelligent control system.

# 1 Introduction

Unmanned surface vehicles (USVs), which have the characteristics of high speed, safety and versatility, are a new typical application field of ship intelligence in hazardous ocean surface environments (Liu et al., 2016; Campbell et al., 2012). Surface vehicles with autonomy developed and demonstrated by academic labs, corporations and government users, have a clear advantage over remote controlled vehicles (Braunl et al., 2006; Manley, 2008). Today, USVs have a wide range of applications, ranging from port monitoring, to hydrographic surveys, and even to maritime search and rescue.

Unmanned surface vehicles cover a wide range of technologies, including traditional ship technology, multi-sensor intelligent monitoring technology, automatic obstacle avoidance technology, high reliability and high redundancy data transmission technology, automatic fault detection technology for electromechanical, and autonomous navigation technology (MUNIN Project, 2016; Liu et al., 2016; Campbell et al., 2012). At present, automatic path planning technology, water surface objects detection and target automatic identification technology, autonomous decision-making and obstacle avoidance technology, and communication technology have become key technologies in the development of USVs (The navy unmanned surface vehicle (USV) master plan, (2007), p.67).

The future progress of USVs depends on the development of full-autonomy, enabling USVs to work in any unstructured or un-predictable environment without human supervision (Liu et al.,2016; Campbell et al., 2012). So, the basic problem in the development of an autonomous system for USV lies in the construction of a reliable obstacle avoidance system, which is mainly determined by planning a path to avoid static obstacles and dynamic obstacles when USV sails in complex environments (Colito, 2007; Tang et al., 2015; Kim et al., 2014). Path planning problem, as the fundamental aspect of USV guidance system, can be divided into global path planning and local path planning. Global path planning is the basic problem that needs to be solved firstly in unmanned guidance system, so this paper mainly focuses on this problem.

Many researchers make efforts to solve the classic problem of searching the optimal global path in a cluttered environment over the years. Dijkstra (1959) proposed an algorithm to solve the shortest path problem. Although Dijkstra algorithm can find the optimal path, it searches for nodes in every direction without heuristic, which makes this algorithm inefficient and unreasonable in practical applications. Hart et al. (1968) described how heuristic information from a problem domain can be incorporated into a formal mathematical theory of graph searching and proposed the $A^*$ search algorithm, which can quickly find an optimal path with the least number of nodes. Liu et al. (2017) recently proposed an improved $A^*$ path-searching algorithm considering the vehicle powertrain and fuel economy performance. This algorithm solves the problem of global path planning for autonomous vehicles in off-road environment on the basis of the DEM map. In the field of global path planning for USV, noteworthy researches have been carried out to solve static obstacles avoidance. One such paper (Yang et al., 2015) converts satellite thermal images into binary images, and uses an advanced $A^*$ algorithm to determine safer and suboptimal paths for USVs. Binary images contain incomprehensive information and the analytical results of satellite thermal images cannot be reused sometimes, so this algorithm is not suitable for large-scale global path planning problems. Zhuang et al. (2011) presented a search algorithm for shortcut based on electronic charts. This algorithm reduces planning time and improves planning precision, but it does not solve the inefficient shortcoming of Dijkstra algorithm effectively. Song (2014) introduced an improved ant colony algorithm for global path planning of USVs, and the main idea of this algorithm is distributing each ant's route dynamically. Wang et al. (2017) proposed an improved $A^*$ algorithm for sailing cost optimization considering sailing safety and path smoothing based on electronic charts. The simulation results show the improved $A^*$ algorithm can generate safe and reasonable global path for USVs between two points. Kim et al. (2013) proposed a curvature path planning algorithm for USVs, where both the actual dynamic constraint of

USVs, and the environmental information are explicitly considered in the non-uniform cost map.

This paper mainly solves the global path planning problem abstracted from practical applications among multiple task points, and the research results can be applied directly to actual applications. The overall scheme for global path planning system is illustrated in Figure 1, where we mainly study the contents in the dashed frame. Compared with existing works，we analyse and extract the path planning requirements of USVs among multiple task points in practical applications. Besides, we first introduce Cube coordinate system and hexagonal grids, which has the advantage of validity, safety and rapidity compared with square grids, to build the global environment modelling based on electronic charts, and use the improved $A^*$ algorithm to determine the global path between two points. Furthermore, we present an improved ant colony optimization (ACO) to realize the global path planning system among multiple task points. Finally, we verify and compare the performance of these algorithms in simulation system.

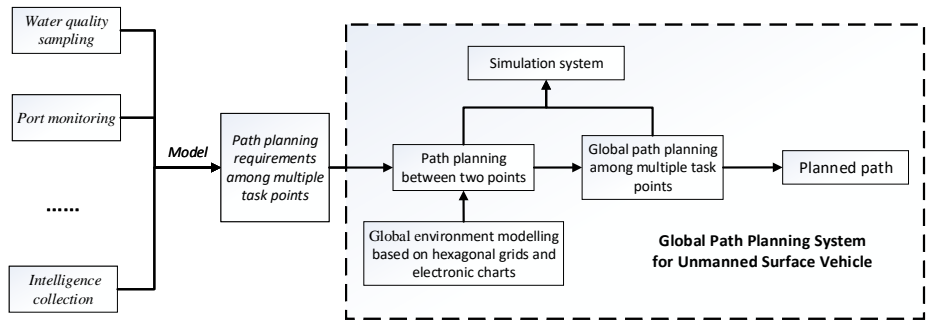

Figure 1. Global path planning system architecture of multiple task points

## 2  Global environment modeling

This paper meshes marine environments by hexagonal grids, which are labelled as unnavigable or navigable according to static obstacles information extracted from electronic charts, and presents the concept of "grid weight" from the perspective of safety. In addition, this section demonstrates that hexagonal grids are superior to traditional grids in terms of validity, safety and rapidity, and proves that the hexagonal grid only needs to access once in $A^*$ algorithm without subsequent correction and comparison in some case.

### 2.1  Electronic charts analysis

When USV navigates in a marine environment, the surrounding environment can be detected by radar, sonar, camera and other sensors in a small area. However, these sensors cannot provide global marine environment information as USV's voyage larger and larger. Electronic charts just can provide detailed and accurate global marine information, so we can use them to obtain marine geographic information.

Electronic charts mainly contain abundant marine features, including submarine terrain, navigation obstacles, navigation signs, port facilities, trends, currents, and even magnetic anomalies, lighthouses and buoys. The obstacles that global environment modelling focused are all included in electronic charts, so electronic charts as global marine information input can meet actual application demand.

The package format of S-57 electronic charts used in this article follows the international standard of ISO/IEC 8211 (Transfer standard for Digital Hydrographic Data, (2000), p.15), so electronic charts cannot be directly used for global environment modelling input. It is necessary to analyse electronic charts' content according to ISO/IEC 8211 international standard protocol and GDAL/OGR library, and extract the contents that global environment modelling needed (Wang et al.,2017).

## 2.2  Hexagonal environment model

After obtaining marine geographic information from electronic charts, the global environment modelling based on hexagonal grids can be described as: set limited movement area $OS$ to be convex polygon in a two-dimensional plane, and a limited number of static obstacles $a_1, a_2, \ldots, a_n$ distributed in it randomly, where $a_i$ occupies one or more grids. This paper establishes the Cartesian coordinate system in $OS$ where we take the upper left corner as the origin point $O$, longitude as $X$ axis and latitude as $Y$ axis.

Hexagonal grids have two typical layouts, horizontal layout and vertical layout. As shown in Figure 2, this paper uses the "even-r" horizontal layout. Set the hexagon's side length to be $size$, then the unit length of $OS$ region in $X$ and $Y$ axis is $e_x = \sqrt{3} size$ and $e_y = \dfrac{3 size}{2}$ respectively. In "even-r" horizontal layout shown in Figure 2, set the upper left corner $O$ to be ($x_o, y_o$), then the relationship between grid number ($i, j$) and centre point ($x_{ci}, y_{cj}$) of the hexagonal grid is expressed as follows：

$$\begin{cases} x_{ci} = x_o + \dfrac{\sqrt{3}}{2} \cdot size + \dfrac{\sqrt{3}}{2} \cdot size \cdot (i \bmod 2) + \sqrt{3} \cdot size \cdot j \\ y_{cj} = y_o - size - 1.5 \cdot size \cdot i \end{cases} \qquad (1)$$

where mod is remainder operation.

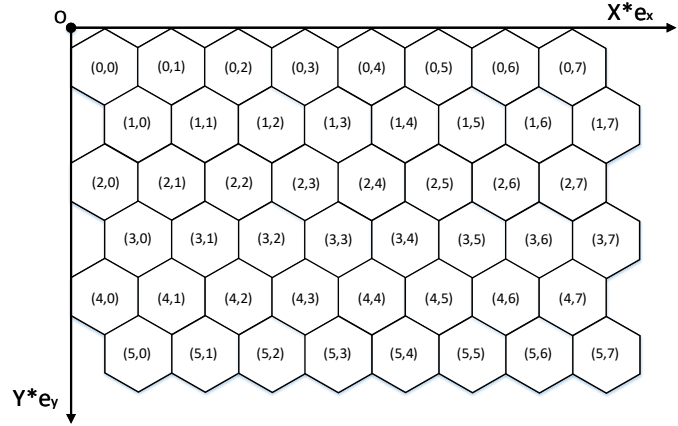

Figure 2. Display of hexagonal grids in "even-r" horizontal layout

The hexagonal environment model of global path planning based on electronic charts and hexagonal grids is shown in Figure 3, where shaded parts represent obstacles.

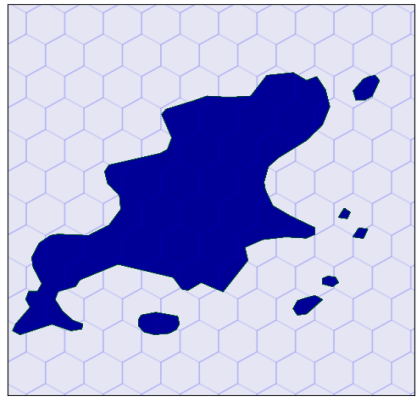

Figure 3. The hexagonal environment model

According to marine geographical information extracted from electronic charts, we could judge whether a hexagonal grid contains obstacles such as mainland, islands, and peninsulas in turn. If there are obstacles in the grid, it is labelled as unnavigable, indicating that USV cannot enter this grid during path planning and sailing. Otherwise, the grid is labelled as navigable.

When USV sails in a navigable grid which near unnavigable grids, it may entry an unnavigable gird because of environmental impact, avoiding obstacle action or motion control error, which seriously threatens the safety of USV. To avoid this potential risk, we propose the concept of "grid weight" to keep USV away from these navigable grids which are adjacent to unnavigable grids. The greater grid weight is, the less possible that USV entries this grid in path planning. Grid weight $w(C)$ of grid $C$ depends on the number of unnavigable grids near it, and its expression, surrounding by $n$ unnavigable grids, is shown in the following equation.

$$grid\ weight \begin{cases} navigable\ grid \begin{cases} w(C) = 1 + \dfrac{n^2}{4} &,\ near\ unnavigable\ grids \\ 1.0 &,\ default \end{cases} \\ null\ ,\ unnavigable\ grid \end{cases} \quad \ldots (2)$$

The hexagonal environment model after navigable labelling is shown in Figure 4, where dark black parts represent unnavigable grids, and the deeper colour in navigable grids means the greater grid weight.

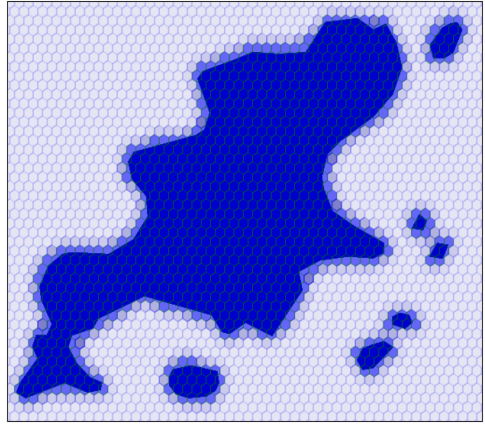

Figure 4. Hexagonal environment model after navigable labelling

Although weighted hexagonal grid map and weighted occupancy grid map can represent whether the grid has been occupied by obstacles, weighted hexagonal grid map mainly reflects potential risk in navigable grids, where the grid weight of navigable grid is determined by the unnavigable grids near it, while weighted occupancy grid map, reflecting the possibility whether obstacles occupy a grid, is mainly to solve the problem of measurement noise when mapping (Konolige, 1997). The greater the weight is, the higher possibility of occupied by obstacles is.

In summary, this paper proposes a method to establish the global environment model of path planning using electronic charts and hexagonal grids. The hexagonal grid in this model contains centre point coordinate, grid size, navigable label, grid weight and other information, which can be directly used for global path planning.

**2.3 Cube coordinate system**

There are a variety of coordinate systems to denote hexagonal positional relationships in hexagonal grids, such as Offset coordinate system, Axial coordinate system and Cube coordinate system (Red Blob Games, 2016). Offset coordinates system, which is similar to Cartesian coordinate system in square grids, is in line with people's cognition, but one of the two axes has to go "against the grain". Compare with Offset coordinates system, Cube coordinate system goes "with the grain" and has simpler algorithms in hexagonal calculation. Besides, Cube and Offset coordinate system can convert back and forth.

There are three primary axes $x$, $y$ and $z$ in Cube coordinate system with a relationship in equation (3) (Red Blob Games, 2016). We use three coordinate values to reflect the geometrical symmetry of grids in Cube coordinate system, which is necessary to handle hexagonal structure (Nagy, 2003). Figure 5 shows the result of this procedure.

$$x + y + z = 0 \tag{3}$$

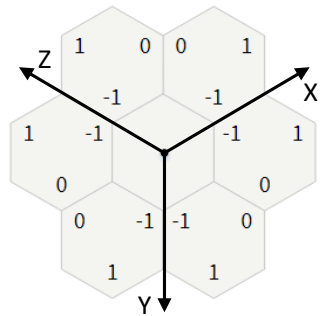

Figure 5. Diagram of Cube coordinate system

Cube coordinate system helps us make hexagonal grids' algorithms simpler. We can reuse standard operations from Cartesian coordinates: e.g. adding coordinates, subtracting coordinates, multiplying, and dividing by a scalar, and distances in Cube coordinate system (Red Blob Games, 2016).

Cube and Offset coordinate system can convert back and forth in hexagonal grids. In "even-r" horizontal layout, set a hexagonal grid to be (col, row) in Offset coordinate system, corresponding to ($x$, $y$, $z$) in Cube coordinate system, the mutual convert formula between Offset and Cube coordinate system can be expressed as:

- convert cube to "even-r" offset：

$$\begin{cases} col = x + (z + (z \& 1))/2 \\ row = z \end{cases} \quad (4)$$

- convert "even-r" offset to cube:

$$\begin{cases} x = col - (row + (row \& 1))/2 \\ z = row \\ y = -x - z \end{cases} \quad (5)$$

According to equation (5) above, hexagonal grids in Figure 2 can be also expressed as Figure 6 in Cube coordinate system.

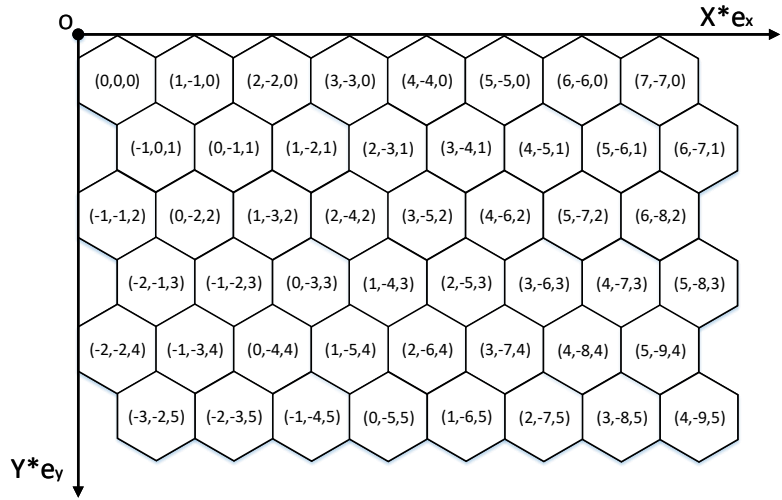

Figure 6. Hexagonal grids layout in Cube coordinate system

Cube coordinate system is more convenient than Offset coordinate system to express the relative algorithms in hexagonal grids, making hexagonal grids algorithms simpler. So, this paper uses Offset coordinates system for map storage and displaying, and introduces Cube coordinates system to express many algorithms in hexagonal grids.

**2.4 Comparison between hexagonal grids and square grids**

The geometric shape of traditional grids is mainly referred to square whose path planning strategy is quadtree or octree. This paper compares hexagonal grids with square grids to prove that hexagonal grids have the advantages of validity, safety and rapidity.

Validity: The square grid extends to four directions in quadtree strategy, so the rotary angle is 90 degrees when changing direction, which results in longer planned path. The hexagonal grid changes its orientation by 60 degrees each time, which reduces step angle significantly and improves the validity of path. The efficiency $\kappa$ (ratio of the straight distance and the distance of planned path) is introduced to judge the planned path. In Figure 7, the planed path is $A \to B \to C$ in quadtree strategy, and its $\kappa$ is 0.707. Similarly, the planed path is $D \to E \to F$ in hexagonal grids, and its $\kappa$ is 0.866. As we can see, the planed path in hexagonal grids is shorter than square grids.

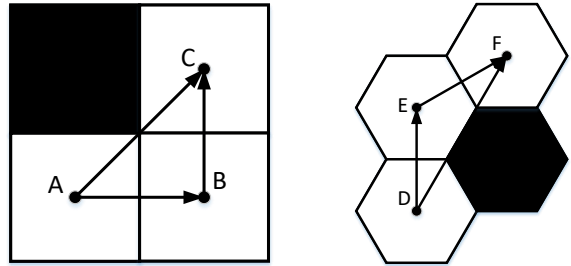

Figure 7. Comparison of square grids and hexagonal grids

Safety: The square grid extends to eight directions with 45 degrees in octree strategy, which is easy to encounter obstacles, so the planed path in octree strategy is usually dangerous and relatively conservative. As shown in Figure 7, where the black grid represents unnavigable grid, the shortest distance between the path $A \to C$ and unnavigable grids is 0 unit in square grids, while the shortest distance between the path $D \to E \to F$ and unnavigable grids is 0.5 unit in hexagonal grids.

Rapidity: Provided that the hexagonal grid $i$ and $j$ access to the same grid $k$, and $i$ is one of the traceable parent grids of $j$, then we prove below that the evaluation value $f(k_i) < f(k_j)$ in $A^*$ algorithm. It means that the hexagonal grid only needs to access once in such case, and subsequent access is superfluous, which can reduce computational complexity and speed up calculation of $A^*$ algorithm significantly.

As shown in Figure 8 (a), the evaluation values of grid $A$, $B$ and $C$ are $f(A)$, $f(B)$, $f(C)$. When $A$ accesses to adjacent $B$ and $C$, it extends to $B$ firstly if $f(B) < f(C)$, then $B$ will access to peripheral grids, which makes $C$ accessed again. $f(C_1)$ is the evaluation value of $C$ when the extended path is $A \to C$, while $f(C_2)$ is the evaluation value of $C$ when the extended path is $A \to B \to C$ in Figure 8 (a), and they can be easily constructed as equation (6) and (7). Since the weight of $B$ is constant above 0, then $\Delta C$ is greater than 0 in equation (8), so we can prove that $f(C_1) < f(C_2)$.

$$f(C_1) = g(A) + w(C) + h(C) \tag{6}$$

$$f(C_2) = g(A) + w(B) + w(C) + h(C) \tag{7}$$

$$\Delta C = f(C_2) - f(C_1) = w(B) > 0 \tag{8}$$

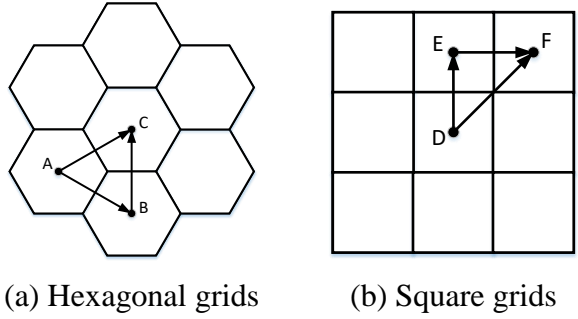

(a) Hexagonal grids      (b) Square grids

Figure 8. Extension diagram of $A^*$ algorithm

However, the traditional square grids cannot guarantee that. As shown in Figure 8 (b), the extension procedure of square grids is similar to hexagonal grids. In square grids, $f(F_1)$ is the evaluation value of $F$ when the extended path is $D \rightarrow F$ in equation (9), and $f(F_2)$ is the evaluation value of $F$ when the extended path is $D \rightarrow E \rightarrow F$ in equation (10). Grid weight $w(E)$ and $w(F)$ only depends on grid $E$ and $F$, and the sign of $\Delta F$ is uncertain, so the calculation, comparison and even correction operations for $f(F)$ is necessary when grid $F$ is accessed again.

$$f(F_1) = g(D) + \sqrt{2} \cdot w(F) + h(F) \tag{9}$$

$$f(F_2) = g(D) + w(E) + w(F) + h(F) \tag{10}$$

$$\Delta F = f(F_2) - f(F_1) = w(E) - (\sqrt{2} - 1) \cdot w(F) \tag{11}$$

In addition, hexagonal grids are more suitable for ocean and other open environments. The area ratio of inscribed circle to square is 0.7854, while the value is 0.9069 in hexagon. Especially, the two-dimensional projection of most obstacles in oceans is more likely to be circle rather than square, so the coverage of hexagonal grids on obstacles is significantly higher than square grids.

In summary, hexagonal grids solve the limitation of traditional square grids which cannot balance security and effectiveness at the same time, improve path search speed, and increase the safety of planned path.

# 3 Algorithm design of global path planning system

USV has a wide range of applications in civilian domains, mainly including maritime safety, port monitoring, hydrological surveying, water quality sampling, maritime search and rescue (MUNIN Project, 2016; Manley, 2008). Combine with the requirements of port monitoring, water quality sampling, hydrological survey and other tasks, this paper establishes the demand model of global path planning for USV.

This paper mainly researches the touring problem of multiple task points, and designs the high security and reasonable path that USV starts from starting point, and tours multiple task points to complete special tasks, then returns to target point. For example, in water quality sampling application, global path planning system plans the path that USV tours various sampling points, completing the automatic collection of water samples in each sampling point, and returns to target point for water quality analysis. Therefore, this path planning problem has a strong application background, and the planned path can be directly applied to specific applications.

The global path planning system of USV among multiple task points can be divided into two parts in local and global perspectives, one part is the path planning between two points, and the other is searching the optimal path which tours multiple task points based on the planned path between two points.

## 3.1 Path planning between two points

The goal of path planning between the starting and target point is to find the optimal path, which avoids environment obstacles autonomously and safely. At present, there are Dijkstra, $A^*$ algorithm, artificial potential field method, genetic algorithm, ant colony algorithm in path planning research field (Zhang et al., 2003). These algorithms have their own advantages and limitations. According to the application type and task requirement, this paper chooses $A^*$ algorithm, which is more suitable for global path planning in grids, to search the optimal path. Hart (1968) utilized the heuristic information into a formal mathematical theory of graph searching and combined the Dijkstra algorithm with the heuristic information to form a new searching strategy, which is known as $A^*$ algorithm.

### 3.1.1 $A^*$ algorithm

$A^*$ algorithm is a classical heuristic search algorithm in global path planning. The basic theory of $A^*$ algorithm is to calculate the evaluation value of each adjacent child nodes that current node may reach. When the path search process extends to node $i$, node $i$ needs to be compared with all nodes that have been extended. If node $i$ is an extended node, it indicates that a new path to node $i$ has been found. If the new path makes the evaluation value $f(i)$ smaller, it is necessary to modify node $i$ that it points to a new parent node of the new path. Finally, this algorithm selects the minimum evaluation node to expand sequentially and put it into the closed table. Repeat these previous steps until the target point is reached or the open table is empty (De Smith et al., 2007).

### 3.1.2 The improved A$^*$ algorithm

Under the security constraint, the optimal goal of global path planning for USV is no longer the distance optimization, but sailing cost optimization. This paper proposes an improved A$^*$ algorithm to find the optimal path with the minimum sailing cost between two points.

The real distance in hexagonal grids is half distance in Cube coordination system (Nagy, 2003). The Manhattan distances $D(u,v)$ between $u$ and $v$ in Cube coordinate system can be expressed as：

$$D(u,v) = \frac{|u.x - v.x| + |u.y - v.y| + |u.z - v.z|}{2} \tag{12}$$

The sailing cost $G(u,v)$ from $u$ to adjacent $v$ is the product of the Manhattan distance $D(u,v)$ and the grid weight $w(v)$ of $v$ in equation (13). We can see from equation (13) that these navigable grids near obstacles or shallow water areas have a higher sailing cost, so the planned path will be far away from these potentially dangerous navigable grids.

$$G(u,v) = D(u,v) \cdot w(v) \tag{13}$$

There may be multiple grids with the same evaluation value searched in path planning algorithm, but we only want to choose the path closing to the line from starting point to target point, whereas the traditional A$^*$ algorithm cannot do that. This paper introduces the "guidance value" to modify heuristic function in equation (17), which can reduce the number of grids with same evaluation value and choose the optimal path we expected.

Provided that the line from grid $i$ to target point is $L_1$, and the line from starting point to target point is $L_2$. We can obtain the sine angle $\sin\theta$ of deviation between $L_1$ and $L_2$, and the guidance value $p(i)$ of grid $i$ can be expressed as:

$$p(i) = 3/(4 - \sin\theta) \tag{14}$$

In this paper, we define the evaluation function $f(i)$ in the improved $A^*$ algorithm as:

$$f(i) = g(i) + h(i) \quad (15)$$

where $g(i)$ is the sailing cost from starting point $S_0$ to node $i$, $h(i)$ is the heuristic function of node $i$ to target point $G_0$. Assuming that the planned path from starting point to node $i$ is $S_0 \to n_0 \to n_1 \to \cdots \to n_i$, then $g(i)$ and $h(i)$ can be expressed as:

$$g(i) = \sum_{j=S_0}^{i} G(j, j+1) = \sum_{j=S_0}^{i} (D(j, j+1) \cdot w(j+1)) \quad (16)$$

$$h(i) = D(i, G_0) \cdot p(i) \leq \min\left(\sum_{j=i}^{G_0} (D(j, j+1) \cdot w(j+1))\right) \quad (17)$$

Because the number of hexagonal grids is limited, so the problem of searching the optimal path belongs to the finite graph problem. Hart (1968) gave the following theorem as below:

If $h(n) \leq h^*(n)$ for all n, then $A^*$ is admissible.

where $h^*(n)$ is the actual cost of an optimal path from $n$ to a preferred goal node $s$, $h(n)$ is the heuristic function of node $n$ to target node.

In this paper, $h(i)$ defined in equation (17) is less than or equal to the Manhattan distance between node $n$ and target node, $h^*(i)$ is defined as below:

$$h^*(i) = \min\left(\sum_{j=i}^{G_0} (D(j, j+1) \cdot w(j+1))\right) \geq D(i, G_0) \geq h(i) \quad (18)$$

where heuristic function $h(i)$ is the lower bound of $h^*(i)$, so it guarantees that the improved $A^*$ algorithm can find the optimal solution in global path planning.

### 3.1.3 Path smoothing

There are ladder or jagged lines in the planned path based on the improved $A^*$ algorithm sometimes, and there may be more polylines, more turning times and more turning angles because of extra waypoints in practice, so we propose a smoothing method to remove extra waypoints without reducing the safety of the planned path. The smoothing method can make folding lines into straight lines and increase the smoothness of planned path.

In this paper, the smoothing method will traverse all nodes in the path. Firstly, we consider 3 consecutive nodes $i$, $i+1$ and $i+2$ (the initial value of $i$ refers to the starting point). When there is no obstacles in the connection-region between node $i$ and $i+2$, and the safety of segment $i \rightarrow i+2$ is not lower than the previous segment $i \rightarrow i+1 \rightarrow i+2$, then we can remove the redundant $i+1$. Repeat these previous steps until the connection-region between node $i$ and $j$ through obstacles. Secondly, we consider 3 consecutive nodes $j-1$, $j$ and $j+1$ after node $i$. Repeat the above steps until we have traversed all nodes in the path.

In practical applications, because of control precision and positioning error of USV, it is hard to guarantee that USV can accurately reach the definite waypoint. So, we put forward the concept of "arrived radius", which defines a certain region near the waypoint, and we consider that USV has arrived at the waypoint as long as USV reaches it. Once USV arrives at a waypoint, it will turn towards the next waypoint of the planned path, which relates to the problem of turning.

Considering the practical issues in the process of turning, e.g. the minimum turning radius of USV and the unwanted sharp turns, we design two kinds of turning algorithm in Figure 9 according to turning angle and the minimum turning radius. When turning angle is large, the situation is shown as Figure 9 (a), when turning angle is small, the situation, which is rare in practical, is shown as Figure 9 (b), and the critical angle $C_\theta$ is

$$C_\theta = 2 \cdot \arctan(\frac{minimum\ turning\ radius}{arrived\ radius}) \quad \dots\dots\dots\dots(19)$$

The turning center and turning radius can be calculated from these situations. In practical application, each waypoint of the planned path not only contains coordinate information, arrived radius, also includes turning center and turning radius. These parameters, which meet dynamics characteristics of USV, can guarantee a smooth path following or trajectory tracking performance of USV.

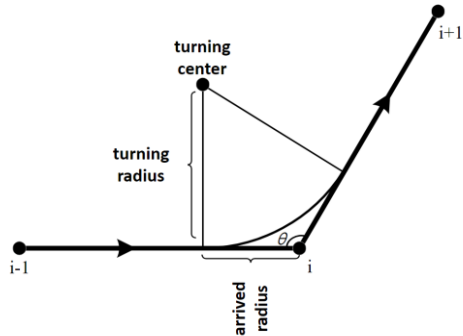

(a) Turning inside situation

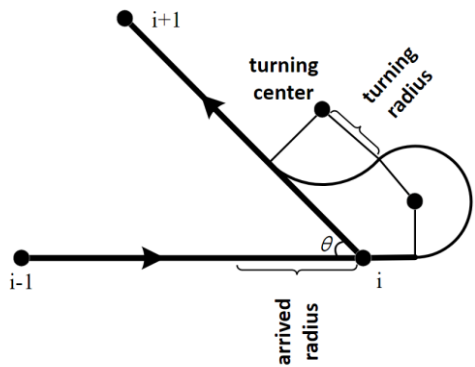

(b) Turning outside situation

Figure 9. Two kinds of turning situation

This paper introduces guidance value and path smoothing method, and considers practical issues in global path planning. Furthermore, we propose an improved A$^*$ algorithm based on sailing cost optimization, this algorithm can find a safer global path with minimum sailing cost between two points quickly and effectively.

### 3.2 Path planning among multiple task points

In the global path planning for USV, the previous researchers mainly focus on solving the global path planning problem of two points without considering how to apply it to specific applications, that directly leads to limited applications of these research results. Based on the planned path between two points, this paper abstracts the general requirements of global path planning from some applications and proposes an improved ant colony algorithm to solve the global path planning problem among multiple task points.

### 3.2.1 Multiple task points modelling

This paper analyses and extracts the global path planning requirements of port monitoring, water quality sampling, hydrological survey and other tasks for USVs. To sum up, multiple task points are given in a complex environment with obstacles, where we should design the global path allowing USV to tour each task point safely with minimum sailing cost.

The path planning modelling of multiple task points can be abstracted as the shortest path problem of network in Figure 10. In the abstract mathematical view, the network is a directed graph with weight essentially, and it consists of nodes, directions and arcs. In the paper, nodes refer to starting point, target point and task points, and arcs refer to the planned path between nodes, and sailing cost between points can be used as the weight of arcs, so the goal of path planning among multiple task points is to find the optimal path with the minimum distance in the network.

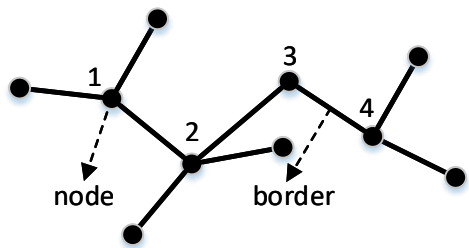

Figure 10. A network diagram consists of nodes and borders

### 3.2.2 Ant colony optimization

Ant colony optimization (ACO) is a popular metaheuristic that can be used to find approximate solutions of difficult optimization problems. In ACO, artificial ants search for a good solution to a given optimization problem. The ants incrementally build solutions by moving on a weighted graph. The solution construction process is stochastic and biased by a pheromone model, that is, a set of parameters associated with graph components whose values are modified at runtime by the ants. This procedure is repeatedly applied until a termination criterion is satisfied (Marco Dorigo, 2007).

### 3.2.3 The improved ACO

The global path planning among multiple task points can be abstracted as finding a closed tour that visits each task point once with the minimal length, and ACO is usually used to solve the shortest loop in network graph. To apply ACO, the path planning problem is transformed into the problem of finding the best path on a weighted network graph, but the closed loop cannot be formed if starting point and target point is not the same point in this problem, so we build a virtual border between starting point and target point to form a closed loop in Figure 11, and the weight of virtual border is less than other weights in the network.

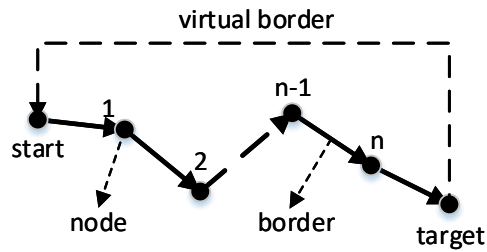

Figure 11. The closed loop of multiple task points

This paper uses ACO to solve the path planning problem, but the basic ACO has the limitation of slow convergence and falling into local optimal solution easily. In this paper, the pheromone and evaporation rate functions are improved by dynamic adjustment of pheromone and evaporation factor, which can achieve better results in global path planning.

Dynamic adjustment strategy: In traditional ACO, pheromone distribution is often concentrated in a certain path that most ants only pass through this path, leading to premature phenomenon. In some situations, pheromone distribution is dispersed to various paths that ants take a relatively long time to search the optimal path, which slows down convergence rate. In this paper, the dynamic adjustment strategy of pheromone and evaporation rate is used to make pheromone distribution relatively uniform, so that the improved ACO can jump out of local optimal solution. In addition, the evaporation rate $\rho$ is directly related to the global search ability of ACO and convergence speed, so the dynamic adjustment of $\rho$ has obvious advantages, it can not only speed up convergence speed, but also improve the search quality (Shang et al.,2009).

The improved ACO can be described as below:

- $C$ is a search space defined over a set of task points.

- $d_{ij}$ $(i, j = 1, 2, ..., n)$ is the distance between task point $i$ and $j$.

- $tabu_k$ $(k = 1, 2, ..., n)$ records the task points that ant $k$ has walked through currently.

$p_{ij}^k(t)$ represents the state transition probability of ant $k$ from $i$ to $j$ at time $t$, and its expression is

$$p_{ij}^k(t) = \begin{cases} \dfrac{\left[\tau_{ij}(t)\right]^{\alpha} \cdot \left[\eta_{ij}(t)\right]^{\beta}}{\sum\limits_{s \in allowed_k} \left[\tau_{is}(t)\right]^{\alpha} \cdot \left[\eta_{is}(t)\right]^{\beta}}, & if \ j \in allowed_k \\ 0, & others \end{cases} \quad \ldots\ldots(20)$$

where $allowed_k = \{C - tabu_k\}$ indicates task points that ant k can select in next step; $\tau_{ij}(t)$ is the pheromone value between $i$ and $j$ at time $t$; $\eta_{ij}(t)$ is the heuristic value associated with the distance $d_{ij}$, and its expression is $\eta_{ij}(t) = \dfrac{1}{d_{ij}}$; $\alpha$ and $\beta$ are positive real parameters whose values determine the relative importance of pheromone versus heuristic information (Marco Dorigo, 2007).

The time-varying function $Q(t)$ of improved ACO in equation (21) replaces the constant pheromone intensity in the Ant-Cycle model as below.

$$Q(t) = \begin{cases} Q_1, & t \leq T_1 \\ Q_2, & T_1 < t \leq T_2 \\ Q_3, & T_2 < t \leq T_3 \end{cases} \quad (21)$$

$\Delta \tau_{ij}^k(t)$ expresses the pheromone left by ant k in path $i \to j$ cycle, which is defined as below:

$$\Delta \tau_{ij}^k(t) = \dfrac{Q(t)}{L_k} \quad (22)$$

where $L_k$ is the total length of the path that ant k has walked this cycle.

To avoid excessive residual pheromone flooding heuristic information, it is necessary to update the residual information when ants tour all task points according to the following equation:

$$\tau_{ij}(t+1) = (1 - \rho(t)) \cdot \tau_{ij}(t) + \Delta \tau_{ij}(t) \quad (23)$$

$$\Delta \tau_{ij}(t) = \sum_{k=1}^{m} \Delta \tau_{ij}^k(t) \quad (24)$$

where $\rho \in (0,1]$ is the evaporation rate, then $1-\rho$ is the residual pheromone factor; $m$ is the number of ants (Marco Dorigo, 2007).

In the improved ACO, the initial value of evaporation rate $\rho$ is $\rho(t_0) = 0.5$. When the optimal solution obtained from this algorithm has not obvious improvement in N cycles, $\rho(t)$ is adjusted as follows:

$$\rho(t) = \begin{cases} 0.98\rho(t-1), & 0.98\rho(t-1) \geq \rho_{\min} \\ \rho_{\min}, & other \end{cases} \quad (25)$$

where $\rho_{\min}$ is the minimum value of $\rho$.

The improved ACO with dynamic pheromone and evaporation rate can get a good balance among accelerating convergence, preventing premature and stagnant phenomena.

## 4 Simulation results

To verify the effectiveness of the global path planning system among multiple task points and evaluate the performance of algorithms in this system, this paper simulates and verifies the path planning algorithms in the system respectively.

The real electronic chart EA200001.000 of South China Sea (The East Asia Hydrographic Commission, 2016) is utilized as the input of path planning modelling to meet the actual situation. The side length of hexagonal grids is 0.002 degree in accordance with USV's kinematic performance. This paper sets starting point, target point and 8 task points in South China Sea region (21.50° N ~ 22.00° N, 112.50° E ~ 113.00° E) in Table 1.

**Table 1** The coordinates of multiple task points in the simulation

| Point Type | Longitude (°) | Latitude (°) |
|---|---|---|
| starting point | 112.663000 | 21.830000 |
| target point | 112.821800 | 21.925600 |
| task point 1 | 112.573625 | 21.714490 |
| task point 2 | 112.620825 | 21.596423 |
| task point 3 | 112.711412 | 21.753520 |
| task point 4 | 112.753098 | 21.664787 |
| task point 5 | 112.877893 | 21.743927 |
| task point 6 | 112.722245 | 21.876427 |
| task point 7 | 112.549900 | 21.655737 |
| task point 8 | 112.849043 | 21.810088 |

### 4.1 Global path planning between two points

In the simulation, this paper sets point (112.663000° E, 21.830000° N) and point (112.753098° E, 21.664787° N) in Table 1 as starting point and target point of USV. We search the planned path with different algorithms and compare the performance of these algorithms.

The simulation results are given in Figure 12. Figure 12 (a) is the result of $A^*$ algorithm in traditional square grids, and Figure 12 (b) is the result of $A^*$ algorithm in hexagonal grids, and the result of $A^*$ algorithm with smoothing in square grids is shown in Figure 12 (c), and the result of the improved $A^*$ algorithm in hexagonal grids is shown in Figure 12 (d), where the solid line means the final planned path, and the dotted line means the path without smoothing.

Table 2 lists the simulation results of several algorithms between two points. In this table, potential hazards represents the total number of unnavigable grids closing to the final path within a grid range, and turning times means the times of changing orientation in planned path, and traversed times represents the times of calculating evaluation value in $A^*$ algorithm, and extended nodes means the total number of nodes which have been visited in $A^*$ algorithm, and the average times which is the ratio of traversed times and extended nodes, represents the average computational times of each extended grid, and the greater it is, the longer computation time is.

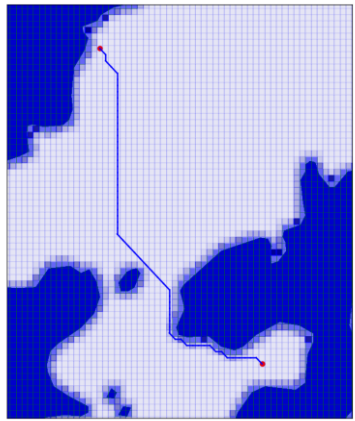

(a) The path planning result of $A^*$ algorithm in square grids

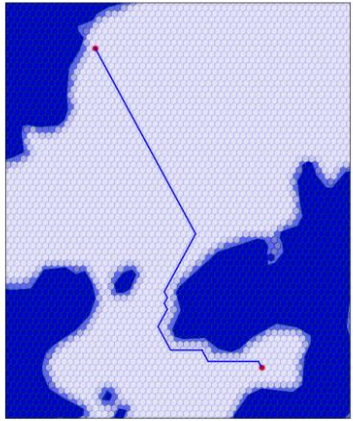

(b) The path planning result of A* algorithm in hexagonal grids

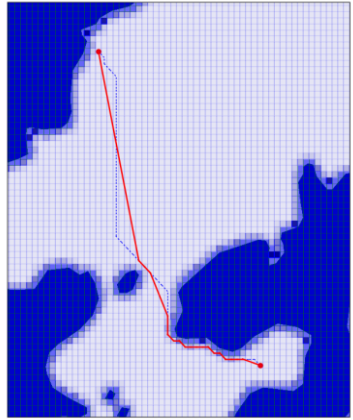

(c) The path planning result of A* algorithm with smoothing in square grids

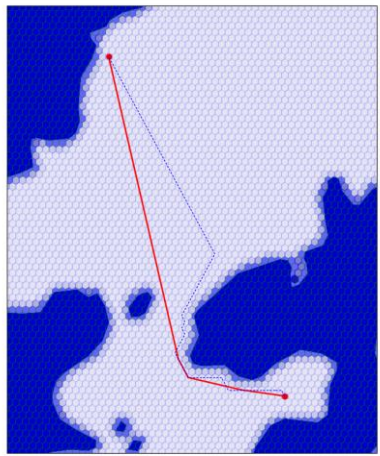

(d) The path planning result of improved A* algorithm in hexagonal grids

Figure 12. Path planning results between two points

**Table 2** Simulation results of several path planning algorithms

|  | traditional A* in square grids | traditional A* in hexagonal grids | A* with smoothing in square grids | improved A* in hexagonal grids |
| --- | --- | --- | --- | --- |
| grid side (degrees) | 0.0032 | 0.002 | 0.0032 | 0.002 |
| distance (nmi) | 14.05 | 14.80 | 13.47 | 13.48 |
| waypoints number | 63 | 69 | 20 | 5 |
| traversed times | 16853 | 1809 | 16853 | 1809 |
| extended nodes | 1413 | 1037 | 1413 | 1037 |
| average times | 11.93 | 1.74 | 11.93 | 1.74 |
| computation time | 11.81s | 5.94s | 12.02s | 6.13s |
| potential hazards | 17 | 0 | 17 | 0 |
| turning times | 14 | 10 | 12 | 3 |

In the simulation results above, the number of hexagonal grids and square grids are equal basically, but their search strategies are different. We can get that hexagonal grids balance security and effectiveness in global path planning from Figure 12 and Table 2, and it is superior to square grids in safety, search speed and traversed times. It can be seen from Figure 12 (b), Figure 12 (d) and Table 2 that the improved A* algorithm can reduce redundant waypoints, improve path smoothness, and shorten sailing distance compared with the traditional A* algorithm. Besides, the path's layout is more reasonable in Figure 12 (d). From Table 2, it can be seen from computation time and average times that the computation speed of hexagonal grids is faster than square grids in global path planning.

It can be seen from the simulation results that the improved A* algorithm can quickly search a safe path avoiding obstacles with relatively minimal navigation distance between two points based on hexagonal grids and electronic charts.

**4.2 Global path planning among multiple task points**

After we complete the path planning and calculate the distance between two points, the planned path and sailing cost can be used as arc connecting two points and weight in the network graph, so the global path planning among multiple task points can be expressed as the network graph in Figure 13.

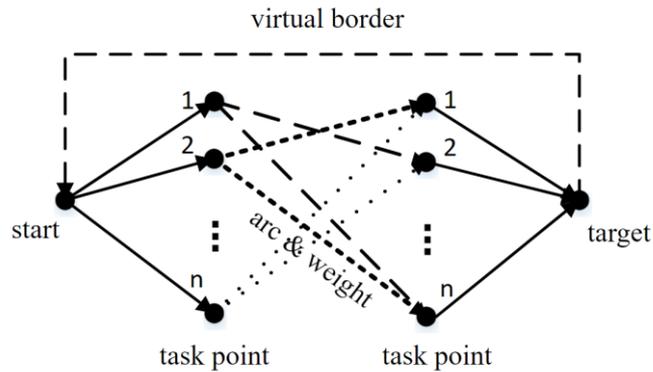

Figure 13. The network graph of multiple task points

The navigability and sailing cost (nautical mile) among multiple task points in the simulation are shown in Table 3, where the sailing cost between two points can be used as the weight of arcs, and the direction between task points is bidirectional.

Table 3. The distances between task points (nautical mile)

|   | $S_0$ | point 1 | point 2 | point 3 | point 4 | point 5 | point 6 | point 7 | point 8 | $G_0$ |
|---|---|---|---|---|---|---|---|---|---|---|
| $S_0$ |   | 9.48 | 16.21 | 5.82 | 13.48 | 15.78 | 4.70 | 13.37 | 11.21 |   |
| point 1 | 9.48 |   | 14.49 | 8.59 | 12.39 | 21.41 | 13.87 | 4.00 | 17.56 | 20.27 |
| point 2 | 16.21 | 14.49 |   | 11.67 | 9.32 | 24.11 | 19.79 | 11.62 | 20.28 | 24.90 |
| point 3 | 5.82 | 8.59 | 11.67 |   | 8.30 | 12.83 | 8.17 | 12.13 | 9.00 | 13.07 |
| point 4 | 13.48 | 12.39 | 9.32 | 8.30 |   | 20.41 | 16.46 | 16.25 | 16.57 | 21.23 |
| point 5 | 15.78 | 21.41 | 24.11 | 12.83 | 20.41 |   | 13.86 | 24.96 | 5.16 | 12.58 |
| point 6 | 4.70 | 13.87 | 19.79 | 8.17 | 16.46 | 13.86 |   | 17.77 | 8.80 | 6.78 |
| point 7 | 13.37 | 4.00 | 11.62 | 12.13 | 16.25 | 24.96 | 17.77 |   | 21.14 | 24.10 |
| point 8 | 11.21 | 17.56 | 20.28 | 9.00 | 16.57 | 5.16 | 8.80 | 21.14 |   | 7.73 |
| $G_0$ |   | 20.27 | 24.90 | 13.07 | 21.23 | 12.58 | 6.78 | 21.14 | 7.73 |   |

The optimal planned path touring multiple task points with the improved ACO is shown in Figure 14, where the value of $Q$ and $\rho$ are dynamically modified as equation (21) and (25) at runtime, and the convergence curve of optimal path is shown in Figure 15. From the simulation, it can be seen that no matter how complicated the environment is, as long as there is a feasible solution among multiple task points, we can find the optimal path with the improved ACO accurately and quickly. In Figure 14, the optimal path departing from starting point is

$$start \to 6 \to 3 \to 1 \to 7 \to 2 \to 4 \to 5 \to 8 \to t\arg et$$

, and the minimal distance is 82 nautical miles in the simulation.

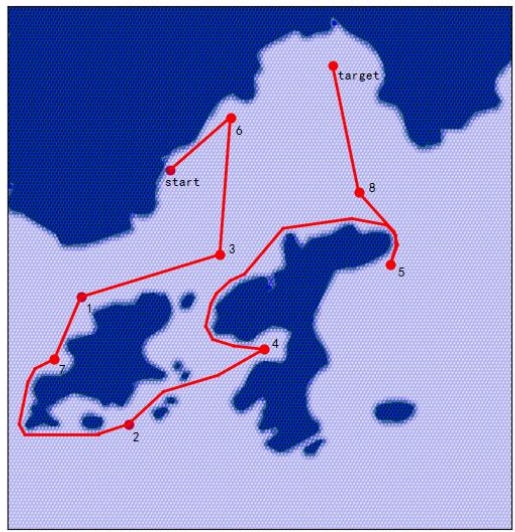

Figure 14. The optimal path among task points in the simulation

It can prove that the improved ACO is superior to traditional ACO in the average optimal path length, the average consumed time, and the average iterations (Shang, 2009).

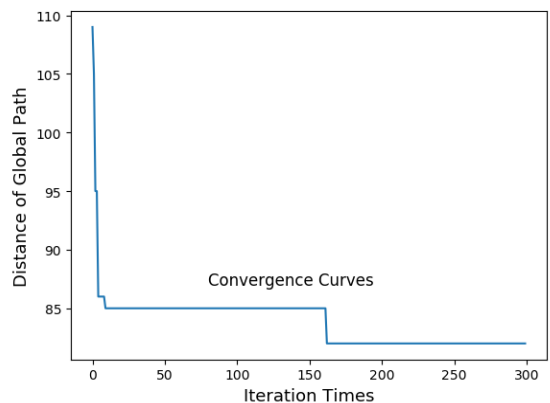

Figure 15. The diagram of optimal path convergence curve

From the simulation results above, we can see that the improved ACO can plan out the global path among multiple task points. It shows a good performance in accelerating convergence, preventing premature and stagnant phenomena, and the navigation distance is relatively minimal.

Therefore, in the environment with known static obstacles, the global path planning system can safely, quickly, and accurately plan the optimal path for USV and produce satisfactory results.

## 5 Conclusion

This paper presents the global environment modelling for marine environment based on hexagonal grids and electronic charts, and proves that hexagonal grids are superior to traditional square grids in validity, safety and rapidity. The improved $A^*$ algorithm based on sailing cost optimization can generate a safe path between two points quickly and accurately.

Furthermore, the improved ACO, which balances quick convergence and preventing premature, realizes the optimal global path planning among multiple task points. Compared with other algorithms in simulation, the global path planning system can safely, quickly and accurately plan a satisfactory path, which is shorter, safer and smoother.

The global path planning problem directly originates from the practical applications, and the global environment modelling is based on real electronic charts, so the path planning module can be called by the motion control system of USV directly, which ensures the practicability and mobility of USV.

The global environment modelling mainly considers the obstacles in electronic charts without considering winds, waves and streams in the marine environment, so this limitation needs to be further researched.